\documentclass[11pt]{article}
\usepackage[preprint]{acl}
\usepackage{times}
\usepackage{latexsym}
\usepackage[T1]{fontenc}
\usepackage[utf8]{inputenc}
\usepackage{microtype}
\usepackage{inconsolata}
\usepackage{amsmath,amssymb}
\usepackage{booktabs}
\usepackage{graphicx}

\title{LLM Evolutionary Search Strategies Should be Compared by Tokens}

\author{
  Tal Oved\thanks{Corresponding author: \texttt{Tal.Oved@ibm.com}} \quad
  Roi Pony \quad
  Oshri Naparstek \quad
  Udi Barzelay \\
  IBM Research
}

\begin{document}
\maketitle

\begin{abstract}
LLM evolutionary search is currently evaluated by counting proposals (candidate programs). We demonstrate this metric hides the true cost of search, and redo the comparison using tokens. By replaying 600 logged runs across 5 tasks and 3 methods, we show that proposal costs vary wildly from 16k to 39k tokens. In equal-proposal comparisons, this silently grants some methods up to $1.9\times$ the compute of others, and gives evolution roughly $6\times$ the budget of independent sampling. When we match token budgets, giving sampling six times more attempts, and remove scorer exploits, the evolutionary advantage largely vanishes. Evolution beats independent sampling by $5\%$ or less on $12$ of $15$ task-method pairs, retaining a $23$ to $28\%$ margin on only $3$, all on one task. A replication with Claude Haiku 4.5 reproduces this split, though a third backbone shows its size depends on the model too. The performance margin justifying these systems is often an accounting artifact. Reporting tokens, not proposals and not model calls, fixes this and costs nothing, as the logs already hold them. \end{abstract}

\section{Introduction}

Large language models now serve as the proposal step in evolutionary search over programs. A model
writes a candidate, an evaluator scores it, the best are kept, and the model is asked to edit
them again. The results are strong enough to change what counts as a discovery method. FunSearch
found new mathematical constructions this way \citep{romera2024funsearch}, AlphaEvolve extended the loop to
algorithms and hardware kernels \citep{novikov2025alphaevolve}, and open engines such as OpenEvolve
reproduce it for public use \citep{openevolve2025}. The same loop drives heuristic design
\citep{liu2024eoh, ye2024reevo}, prompt optimization \citep{fernando2023promptbreeder,
agrawal2025gepa}, and sample-efficient variants \citep{lange2025shinka}, and ADRS-Bench collects
systems-optimization tasks for the setting \citep{cheng2025barbarians}.

These runs are reported in proposals, meaning candidate programs: a search is $k$ seeds grown for $t$
iterations, and methods are compared at matched $kt$. \citet{oved2026evolution} showed that how a fixed budget is
split between $k$ and $t$ changes which method appears to win. Our concern is the unit itself. A proposal is
not a fixed amount of compute, because an evolve prompt carries the current program and part of the archive
and both grow as the run proceeds. A later proposal therefore costs more than an earlier one, and a method
that shows the model more context costs more per proposal. Cost-aware controllers already assume this
\citep{zhang2026costada, ishibashi2026harness}, and what is missing is the measurement of what a proposal costs, and of what it does to a published comparison once that comparison is paid for in tokens.

These systems are usually validated against an ablation that removes
evolution and samples programs independently, run at equal proposals. Independent sampling is a serious
competitor at scale \citep{brown2024monkeys, snell2024scaling, beck2026hill}. If one proposal costs several
times what one sample costs, that ablation got a fraction of the compute it should have had, and the reported
margin is partly an artifact of the accounting.

In this work, we measure both sides. We replay 600 logged runs over 5 optimization tasks and 3 search strategies,
re-indexing every best-so-far curve against cumulative tokens. One proposal costs
16k to 39k tokens depending on strategy and task, so on one task an equal-proposal comparison can hand one strategy $1.9\times$ the compute of another. One proposal costs about six times one independent sample. Paid the same number of tokens, so that sampling gets roughly six times more attempts, evolution wins by
5\% or less on 12 of our 15 strategy and task pairs, one of them indistinguishable from parity, and by 23 to 28\% on the other 3.

\paragraph{Validity inspection.} On transaction scheduling the best program scored $5\times$ the
baseline by placing 3 of 300 transactions, and the scorer accepted it. Scorers that can be gamed are a known
problem \citep{liu2023evalplus, aleithan2024swebenchplus, lehman2020surprising, skalse2022defining}, and
taking the maximum over many runs is how you find the exploits \citep{stroebl2024limits,
agarwal2021precipice}. So we re-run the programs on both sides and score them under the same coverage rule.

\paragraph{Contributions.}
\begin{itemize}
\itemsep0.15em
\item A token-level accounting of LLM evolutionary search: what one proposal costs per strategy and per
task, and how far that is from one independent sample (\S\ref{sec:results}).
\item The comparison against independent sampling redone at equal token cost, with exact order statistics
on both sides (\S\ref{sec:results}).
\item A re-execution audit of candidate programs that removes a $5\times$ headline result and erases the ranking it created on that task (\S\ref{sec:results}).
\end{itemize}

\paragraph{Related work.}
\citet{oved2026evolution}, a concurrent work, asks how to split a fixed $kt$ budget
between seeds and depth; it takes the proposal as the unit, orthogonal to what we test.
\citet{zhang2026costada} price search actions inside a controller to spend a token budget well.
They assert the cost spread; we measure it.
\citet{ishibashi2026harness} also fix a token budget and report that deeper reasoning per candidate beats
more candidates, on one task and with evolution on both sides of every comparison.
\citet{gupta2026noharness} vary harness components budget-matched and find no configuration that wins everywhere, with no sampling arm and no program audit.
\citet{beck2026hill} argue archives and frameworks can be dropped in favor of sampling edits to the best
verified solution, though their sampler is conditioned on that incumbent and compute is reported in
wall-clock time, not tokens.
\citet{rajan2026auditing} measures how often a code RL environment accepts an incorrect solution; we measure
the realized exploit inside a search population, where selection amplifies it. Reward hacking and
specification gaming are long-standing concerns \citep{amodei2016concrete, krakovna2020specification,
pan2022effects, gao2022scaling}.

\section{Experimental Setup}
\label{sec:setup}

\paragraph{Engine, models and strategies.}
All runs use the ADRS engine \citep{cheng2025barbarians} with \texttt{gpt-5-mini}; model, objective and
evaluator remain fixed to isolate the search strategy. The three strategies are
\textbf{OpenEvolve} (islands plus archive, population 40; \citealp{openevolve2025}), \textbf{EvoX}
(co-evolves its selection rule; \citealp{liu2026evox}) and \textbf{AdaEvolve} (adaptive variant;
\citealp{cemri2026adaevolve}). Each task-strategy cell runs extensively for 40 independent seeds and
200 iterations, a budget of 8{,}000 proposals per cell, for 600 logged runs. Throughout, a
\emph{run} is one seed and a \emph{cell} is the forty runs of one task-strategy pair. Every run logs one line per model call with its
prompt, completion and reasoning token counts, and one row per evaluated program with its score and a
timestamp, which is what lets us price the search afterwards.

\paragraph{Tasks and baseline.}
We use three system tasks from ADRS-Bench \citep{cheng2025barbarians}: \textbf{PRISM}, an LLM-serving
scheduler measured by goodput; \textbf{Cloudcast}, a broadcast planner measured by transfer cost; and
\textbf{transaction scheduling}, measured by throughput. We add two mathematical construction tasks,
circle packing in a square and the Heilbronn triangle problem, giving five tasks, fifteen cells. Each
task is optimized through its native \texttt{combined\_score}. Against these we run a sampling baseline:
8{,}000 independent programs drawn from the task description alone, with no history or archive, scored
by the same evaluator. This is the control used in prior work \citep{oved2026evolution}; our contribution is to charge
both sides the same way.

\section{Method}
\label{sec:method}

\paragraph{Notation.}
Seed $i$ of a cell produces candidates with scores $f(x_{i,1}),\dots,f(x_{i,t})$, and its value at depth
$t$ is the running best
\begin{equation}
M_i(t)=\max_{1\le s\le t} f(x_{i,s}).
\end{equation}
Running $k$ seeds and keeping the best gives $\mathcal{B}_k(t)=\max_{i\le k} M_i(t)$. Prior work budgets
this by proposals, $B=k\,t$.

\paragraph{Pricing a run.}
The two logs share one clock, so the tokens a seed has spent by the time its $j$-th program appears is the
cumulative sum over the model calls that precede it. Every seed therefore becomes a curve of best score
against tokens spent rather than against iterations. Summing over the calls, and not over the proposals,
matters because a proposal is not one call: retries and the occasional meta-call that a strategy makes to
steer itself are charged here and are invisible in a proposal count.

\paragraph{Comparing at equal cost.}
Given a token budget $B$, we let $k$ seeds share it and read each seed's curve at $B/k$, then take the best
over $k$ and maximize over $k$. Which $k$ of the 40 seeds a practitioner draws is averaged out with exact
order statistics, so no number depends on one ordering and there is no resampling noise. The baseline is
priced the same way: a budget buys $B$ over the cost per sample, and the expected best of that many is again
an exact order statistic. We cap every budget at what the sampling pool covers, between 28 and 48 million
tokens, preventing extrapolation. We report proposal-budget and token-budget comparisons side by side, isolating accounting from search.

\paragraph{Auditing the programs.}
Three of these scorers accept programs that fail the task, and two of them can only be
caught by running the program again, so we re-execute those and check coverage
directly: for transaction scheduling, how many of the 300 transactions a schedule places, and for Cloudcast,
the fraction of required partitions that arrive. Both sides are judged by identical code. Because such a
program can lead midway through a run, we walk each seed in order, keep the best score surviving the
check, and re-execute a candidate only when it would beat it, about six re-runs per seed versus two
hundred. Where logs already record coverage, as PRISM does, we read it from them.

\section{Results}
\label{sec:results}

\paragraph{A proposal is not a unit of compute.}

\begin{figure*}[t]
\centering
\includegraphics[width=\textwidth]{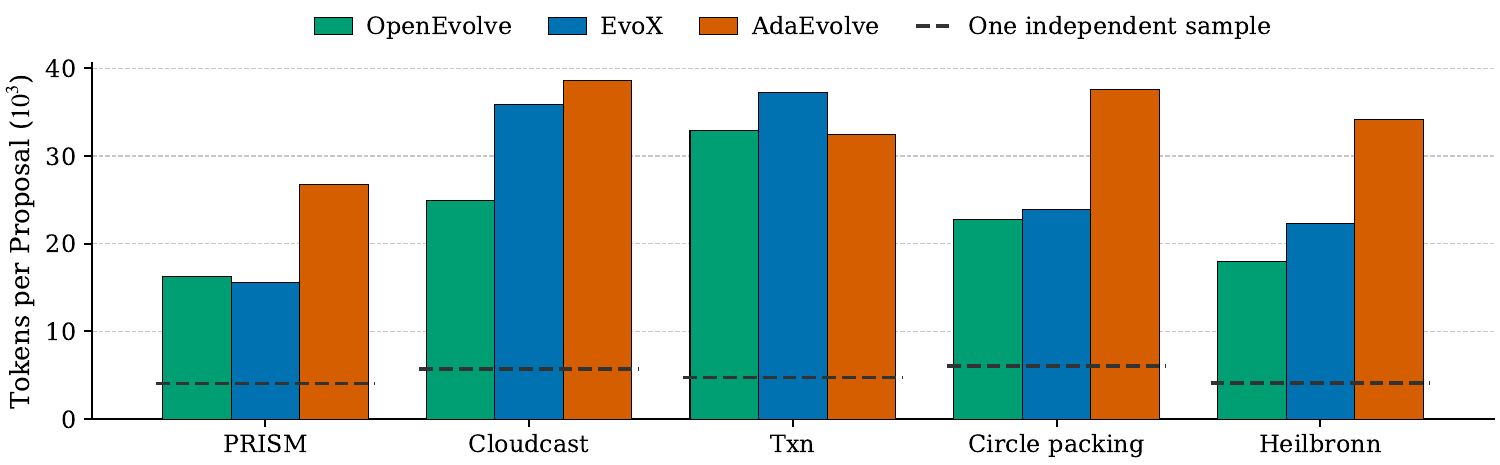}
\caption{Tokens per proposal, by method and task, with the independent sample cost as the dashed
line. Equal-proposal comparisons yield equal compute only if these bars match, which they do not: the same
method varies by up to $2.4\times$ across tasks, and within a task the methods differ by up to
$1.9\times$. The distance to the dashed line ($4.6\times$ to $7.2\times$) is why a sampling baseline run at
equal proposals is given far less compute than the compared methods.}
\label{fig:bars}
\end{figure*}

Figure~\ref{fig:bars} and Table~\ref{tab:census} demonstrate it. One proposal costs 23.0k tokens
under OpenEvolve, 27.0k under EvoX and 33.9k under AdaEvolve, and between 15.6k and 38.6k across cells, a
spread of $2.5\times$. One proposal also takes 1.3 to 1.8 model calls on average, because retries and extra
calls some methods make to steer themselves are not proposals. Reasoning tokens are 5 to 24\% of a
proposal.

Appendix~\ref{app:haiku} repeats this on Haiku 4.5, over two math tasks and all three
strategies at the same 40 seeds by 200 iterations. AdaEvolve remains dearest, where
OpenEvolve is 41\% cheaper per proposal for one task and 9\% dearer for the other.

Two consequences. Across tasks, an equal-proposal comparison gives AdaEvolve
$1.5\times$ OpenEvolve's compute, and in one task reaching $1.9\times$, so part of any
reported advantage is a larger budget. One cell, forty runs by 200 iterations, costs 108--309 million tokens, averaging 194 million, notable in a field reporting
iteration counts.

\begin{table}[t]
\centering\small
\begin{tabular}{lrrr}
\toprule
Method & tokens/prop. & calls/prop. & reasoning \\
\midrule
OpenEvolve & 22,985 & 1.29 & 8\% \\
EvoX       & 26,992 & 1.82 & 9\% \\
AdaEvolve  & 33,907 & 1.52 & 14\% \\
\midrule
one sample & 4,064--6,050 & 1.0 & 24--47\% \\
\bottomrule
\end{tabular}
\caption{Cost of one proposal, averaged over tasks, against the cost of one independent sample. Reasoning
tokens are a share of the total. The sample column is a range because it varies by task.}
\label{tab:census}
\end{table}

\paragraph{Sampling is roughly six times cheaper per attempt.}

An independent sample costs 4.1k to 6.0k tokens, against 16k to 39k for a proposal
(Figure~\ref{fig:bars}). The ratio is
$4.6\times$ on circle packing, $4.8\times$ on PRISM, $5.8\times$ on Cloudcast, $6.0\times$ on Heilbronn and
$7.2\times$ on transaction scheduling, a mean of $5.7\times$. The cause is structural: an evolve prompt carries the program being edited and part of the archive, and both grow as
the run proceeds, while a fresh sample carries only the task description.

This matters since most works in the field, run this baseline at equal proposals. At equal tokens
it should get more attempts than it was given.

\paragraph{At equal cost, evolution wins less.}
Table~\ref{tab:main} gives both sides at two budget definitions, and Figure~\ref{fig:curves} in
Appendix~\ref{app:curves} plots the full curves.
The correction comes first. At equal proposals the largest advantage was transaction
scheduling with AdaEvolve at $5.06\times$; that cell's winner places 3 of 300 transactions. it
reads $5.06$ in both currencies, because its exploit is found early and cheaply, leaving nothing for more
tokens. Auditing all
three methods on that task drops the equal-token ratio from $5.06$ to $\mathbf{1.02}$ for AdaEvolve, from
$1.14$ to $\mathbf{1.02}$ for OpenEvolve, leaving EvoX untouched at $1.04$ because its forty
winners place every transaction.

What remains is small; intervals agree: only Cloudcast with OpenEvolve spans parity, PRISM's three sit just above, and only Heilbronn's exceed $1.17$ (Table~\ref{tab:ci}). The largest advantages are all Heilbronn, $1.23$--$1.28$ across methods, and on PRISM all three land within $1\%$. Moving from equal proposals to equal tokens removes some advantage in
$12$ of $15$ pairs, excepting the three PRISM pairs already at parity, and on
Cloudcast with OpenEvolve removes it all, leaving sampling ahead. Counting model calls instead does not fix it either (Appendix~\ref{app:calls}). Reporting in proposals flatters evolution twice: once by not charging for larger prompts, and
again if nobody checks its best program's output.

The final sample doubling yields $+0.00\%$ on circle packing, $+0.38\%$ on PRISM, $+0.44\%$ on transaction scheduling, $+1.37\%$ on Heilbronn and $+3.63\%$ on Cloudcast. Sampling has saturated here while evolution still rises, so the window favours sampling and reported margins are lower bounds.

\begin{table}[t]
\centering\footnotesize
\setlength{\tabcolsep}{4.5pt}
\begin{tabular}{@{}llrrr@{}}
\toprule
& & \multicolumn{2}{c}{evolve / sampling} & \\
\cmidrule(lr){3-4}
Task & Method & prop. & tok. & cut \\
\midrule
Heilbronn     & EvoX    & 1.33 & \textbf{1.28} & 13\% \\
Heilbronn     & OpenEv. & 1.31 & \textbf{1.27} & 10\% \\
Heilbronn     & AdaEv.  & 1.32 & \textbf{1.23} & 27\% \\
Cloudcast     & AdaEv.  & 1.06 & 1.05 & 27\% \\
Cloudcast     & EvoX    & 1.05 & 1.05 & 7\% \\
Circle        & EvoX    & 1.04 & 1.04 & 2\% \\
Circle        & AdaEv.  & 1.04 & 1.04 & 5\% \\
Txn$^\dagger$ & EvoX    & 1.04 & 1.04 & 20\% \\
Txn$^\dagger$ & OpenEv. & 1.03 & 1.02 & 38\% \\
Txn$^\dagger$ & AdaEv.  & 1.03 & 1.02 & 31\% \\
Circle        & OpenEv. & 1.02 & 1.01 & 44\% \\
PRISM         & AdaEv.  & 1.00 & 1.00 & 0\% \\
PRISM         & EvoX    & 1.00 & 1.00 & 0\% \\
PRISM         & OpenEv. & 1.00 & 1.00 & 0\% \\
Cloudcast     & OpenEv. & 1.02 & \textit{0.99} & $>$100\% \\
\midrule
mean & & 1.09 & 1.07 & \\
\bottomrule
\end{tabular}
\caption{Evolution divided by independent sampling at equal proposals (prop.) and at equal tokens (tok.),
the latter at the largest budget the sampling pool supports. The italic entry has an interval spanning parity. The cut column is how
much advantage above parity disappears between them, $(\text{prop.}-\text{tok.})/(\text{prop.}-1)$;
it exceeds $100\%$ when the advantage does not merely shrink. $\dagger$ audited by re-execution: each seed
is credited only with its best program placing all transactions, and three seeds whose entire
upper archive is the exploit are dropped.}
\label{tab:main}
\end{table}

\paragraph{Scorer exploits, and what removing them costs.}
Reward hacking is well documented; we treat it as a filter, reporting changes (Table~\ref{tab:audit}). Three of five scorers accept programs skipping most of the task, and one accepts a working program that misreports. Two exploits win here, and selection spreads both: in the worst transaction-scheduling seed, the top 69 distinct programs are identical exploits, and over a third of candidates score there. Cloudcast's simulator prices plans from submitted edge data, not the provided network, so plans passing coverage checks are billed a fraction of true cost; re-running reproduces this faithfully, so only graph comparisons catch it. One topped the Cloudcast EvoX cell, dropping it from $1.08$ to $1.05$. PRISM is gated throughout, so degenerate winners never reach our numbers, though an ungated best-of-$k$ would ship one under all three methods. Circle packing and Heilbronn validate outputs, serving as controls.

\section{Conclusion}
\label{sec:conclusion}

Report tokens alongside proposals. The logs already hold them; without them an equal-proposal comparison
can vary by $1.9\times$ in compute between methods on one task and by about $6\times$ from the sampling
baseline used to justify it.

Paid in tokens and audited, evolutionary search beat independent sampling by 5\% or less on 12 of 15 pairs, and by 23--28\% on the other 3, all Heilbronn. The difference was the task rather than the method. Heilbronn has a long improvement path where editing a good program keeps paying; on the other four tasks a
fresh sample is almost as good for a fraction of the cost. So check whether a task rewards search before
choosing a searcher.

Audit both sides, and record audit requirements during the run. One stored coverage number per candidate would
replace every re-execution here.

\section*{Limitations}

Logging and evaluating every API call across 1{,}200 full runs imposes a heavy compute cost, which practically limits studies to fewer tasks and search methods. Therefore, we report fifteen pairs, not a broad distribution. The main comparison uses one backbone, with a second in Appendix~\ref{app:haiku} on the two math tasks, where the same task-dependent split appears, and with a third in Appendix~\ref{app:gemini} on the three system tasks. 

The comparison window is capped by the size of the sampling pool, so evolution is measured using about a fifth of its logged corpus. Its curves are still rising there, so the window favours sampling and our margins are lower bounds. This pool size is partly constrained by task difficulty: Cloudcast, for example, rejects a large portion of independent samples, with $38\%$ of draws returning an error and no score, leaving 4{,}898 usable programs out of 8{,}000. 

Finally, prices are reported as token counts, not dollars, because per-token prices change; a reader can simply multiply by the current rate.

\section*{Ethical considerations}

The primary ethical dimension of this work is computational and environmental cost. LLM evolutionary search demands a substantial energy footprint, requiring tens of thousands of API calls per run. Our methodology mitigates this by analyzing existing logged runs rather than executing new searches, maximizing the utility of already-expended compute. Furthermore, by demonstrating that independent sampling can often match evolutionary search at a fraction of the token budget, our findings encourage more energy-efficient uses of language models. The datasets and tasks evaluate system optimization and mathematics, presenting no risks regarding human subjects, privacy, or demographic bias.

\bibliography{reference}

\appendix

\section{Implementation details}
\label{app:impl}

All experiments run on SkyDiscover \citep{liu2026skydiscover}, an open framework for AI-driven algorithmic discovery, using its strategy implementations, task evaluators, and logging. Building on it keeps our pipeline reproducible.

\section{Score against tokens spent}
\label{app:curves}

\begin{figure*}[t]
\centering
\includegraphics[width=\textwidth]{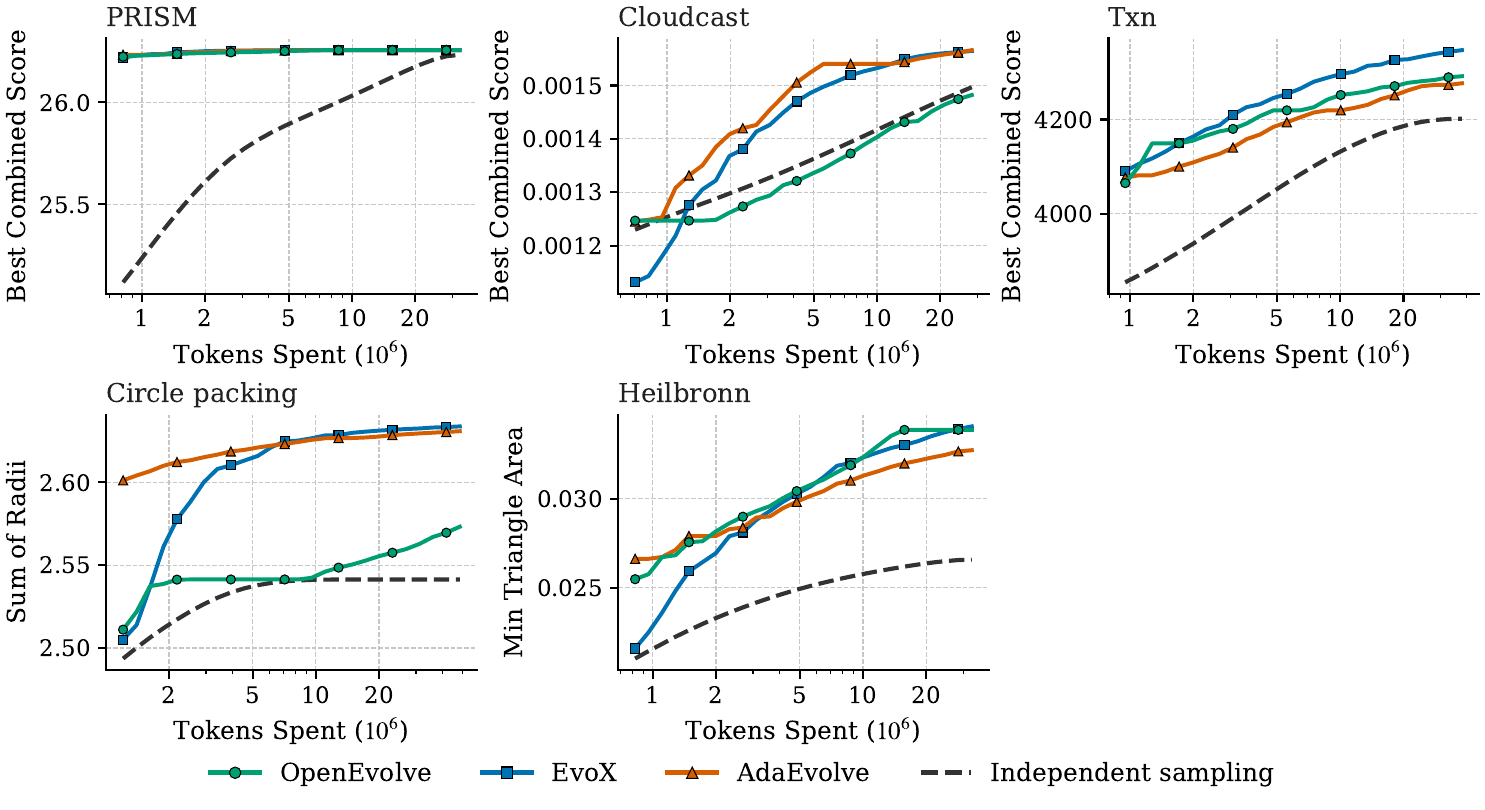}
\caption{Best score against tokens spent. Solid lines are evolutionary search, one per method, each at its
own best number of seeds for that budget. Cloudcast is shown with its coverage gate applied to both sides, and with the two cells whose ceiling misreported its transfer price capped at their best valid program.
The dashed line is independent sampling, which at the same budget gets five to seven times more attempts,
depending on the task. The
transaction scheduling curves are shown after the audit: each seed is capped at the best program that places
all 300 transactions and the three exploit seeds are dropped, so that panel tops out near 4{,}400 rather than
21{,}277. The range stops where the sampling pool stops, so nothing is extrapolated.}
\label{fig:curves}
\end{figure*}

Figure~\ref{fig:curves} plots the curves behind Table~\ref{tab:main}, one panel per task: best score so far
against cumulative tokens, for each search method and for independent sampling on the same axis.

\section{A second backbone ablation}
\label{app:haiku}

\begin{figure}[t]
\centering
\includegraphics[width=\columnwidth]{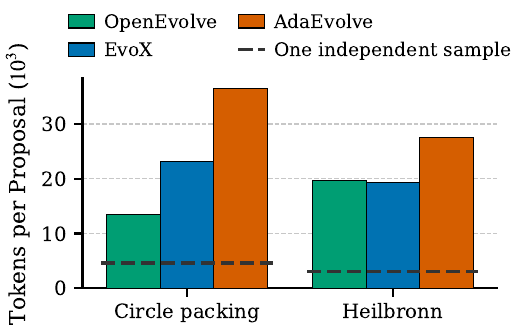}
\caption{Tokens for one proposal on Haiku 4.5, by strategy and task, with the cost of one independent Haiku
sample as the dashed line. All six cells are complete at 40 seeds by 200 iterations.}
\label{fig:haiku}
\end{figure}

We repeat the cost census on Claude Haiku 4.5, using the ablation runs on the two math tasks at the same 40
seeds by 200 iterations, all six cells complete. 
Table~\ref{tab:haiku} and Figure~\ref{fig:haiku} give the result, and the main point is how little changes from the first engine model.
Costs land in the same band on both backbones rather than tracking the model: on circle packing EvoX and
AdaEvolve are within $3\%$ of their gpt-5-mini values, on Heilbronn within $13\%$ and $19\%$, and only
OpenEvolve moves far, $41\%$ cheaper on circle packing and $9\%$ dearer on Heilbronn. What does transfer
cleanly is that AdaEvolve is dearest everywhere, on both tasks and both backbones. The one ordering that does
not survive is OpenEvolve being cheapest: on Haiku Heilbronn EvoX edges it, 19{,}275 against 19{,}645. So the
spread we report in Section~\ref{sec:results} is closer to a property of the harness than of the model,
without being a fixed ranking.

On circle packing one AdaEvolve proposal costs $2.7\times$ one
OpenEvolve proposal, against $1.6\times$ for the same pair on gpt-5-mini. An
equal-proposal comparison is therefore no safer on a cheaper model. One difference worth noting: Haiku reports no reasoning tokens at all, on any of the 58{,}615 calls in these six cells, where gpt-5-mini spends 5 to
$24\%$ of a proposal on them.

We can also repeat the comparison against independent sampling, since we drew a Haiku best-of-$N$ pool for
each task: 8{,}000 attempts, of which 5{,}988 and 1{,}612 produced a valid program, at 4{,}649 and 3{,}076
tokens per attempt. Table~\ref{tab:haiku2} prices both sides the same way as Table~\ref{tab:main}, and the
result carries over. Evolution wins by $1$ to $7\%$ on circle packing and by $17$ to $26\%$ on Heilbronn, so
the task rather than the backbone decides whether the search pays for itself. That is the same split we see on
gpt-5-mini, where Heilbronn returns $1.23$ to $1.28$ and circle packing $1.04$, and it is the clearest evidence
we have that the finding is not an artifact of one model.

\begin{table}[t]
\centering\footnotesize
\setlength{\tabcolsep}{4.5pt}
\begin{tabular}{@{}llrrr@{}}
\toprule
& & \multicolumn{2}{c}{tokens / proposal} & \\
\cmidrule(lr){3-4}
Task & Strategy & gpt-5-mini & Haiku & calls \\
\midrule
Circle    & OpenEv. & 22{,}813 & 13{,}544 & 1.14 \\
Circle    & EvoX    & 23{,}920 & 23{,}177 & 1.61 \\
Circle    & AdaEv.  & 37{,}573 & 36{,}444 & 1.38 \\
Heilbronn & OpenEv. & 17{,}990 & 19{,}645 & 1.84 \\
Heilbronn & EvoX    & 22{,}265 & 19{,}275 & 1.62 \\
Heilbronn & AdaEv.  & 34{,}121 & 27{,}479 & 1.34 \\
\bottomrule
\end{tabular}
\caption{Cost of one proposal on both backbones, averaged over the 40 seeds of each cell.
The calls column is model calls per proposal on Haiku.}
\label{tab:haiku}
\end{table}

\begin{table}[t]
\centering\footnotesize
\setlength{\tabcolsep}{4.5pt}
\begin{tabular}{@{}llrrr@{}}
\toprule
& & \multicolumn{2}{c}{evolve / sampling} & \\
\cmidrule(lr){3-4}
Task & Strategy & prop. & tok. & cut \\
\midrule
Heilbronn & AdaEv.  & 1.34 & \textbf{1.26} & 25\% \\
Heilbronn & EvoX    & 1.31 & \textbf{1.25} & 19\% \\
Heilbronn & OpenEv. & 1.26 & \textbf{1.17} & 35\% \\
Circle    & EvoX    & 1.07 & \textbf{1.07} & 1\% \\
Circle    & AdaEv.  & 1.07 & \textbf{1.06} & 2\% \\
Circle    & OpenEv. & 1.02 & 1.01 & 35\% \\
\midrule
mean & & 1.18 & 1.14 & \\
\bottomrule
\end{tabular}
\caption{Haiku 4.5 evolution divided by Haiku independent sampling: equal proposals (prop.) against equal tokens (tok.), the latter at the largest budget
each pool supports, 37.2M and 24.6M tokens. All six Haiku cells are 40 seeds by 200 iterations. The split is the same one the main text finds on gpt-5-mini,
Heilbronn rewarding search and circle packing barely doing so, which is why we claim it as a property of the
task and not of the model.}
\label{tab:haiku2}
\end{table}

\section{A third backbone: the systems tasks}
\label{app:gemini}

Appendix~\ref{app:haiku} covers the two math tasks on a second backbone. This one covers the three
systems tasks on \texttt{gemini-2.5-flash}, all three strategies and a best-of-$N$ pool each, at the same
forty seeds by 200 iterations. Everything here is gated and audited exactly as in Table~\ref{tab:main}:
PRISM on success rate, Cloudcast on coverage and on declared price, transaction scheduling by
re-execution. That audit found no transaction-scheduling exploit on this backbone, all 118 winners
placing every transaction, and three Cloudcast winners that misreport their price.

Table~\ref{tab:gemini} demonstrates that the accounting problem can be worse. One
proposal costs 16.8k tokens in the cheapest cell and 65.4k in the dearest, a $3.9\times$ spread against
$2.5\times$ on \texttt{gpt-5-mini}, so an equal-proposal comparison misallocates more on this model.
Little of the ordering carries over: AdaEvolve moves from $1.3\times$ to $2.5\times$ the cost of
OpenEvolve, while transaction scheduling with EvoX becomes $29\%$ cheaper rather than dearer. Reasoning
is a third to two thirds of a proposal, against $5$ to $24\%$ before. One independent sample costs $2.1$
to $2.6\times$ more as well, so the gap between a proposal and a sample narrows to $3.3$--$4.4\times$.

Table~\ref{tab:gemini2} gives the comparison. Evolution wins by more on this backbone: the mean
ratio is $1.30$ at equal tokens against $1.07$ on \texttt{gpt-5-mini}, and Cloudcast moves from about
parity to $1.53$--$1.59$. Pricing in tokens still only reduces the advantage, from $1.31$ to $1.30$. What does not hold is
that the size of the margin belongs to the task alone: the same task pays very differently on a different
model. The unit is wrong on both backbones; how much search is worth is not settled by either.

\begin{table*}[t]
\centering\footnotesize
\setlength{\tabcolsep}{5pt}
\begin{tabular}{@{}llrrrrr@{}}
\toprule
& & \multicolumn{3}{c}{tokens per proposal} & & \\
\cmidrule(lr){3-5}
Task & Method & gemini & \texttt{gpt-5-mini} & ratio & calls & reasoning \\
\midrule
PRISM      & OpenEv. & 16{,}849 & 16{,}252 & 1.04 & 1.17 & 43\% \\
PRISM      & EvoX    & 23{,}524 & 15{,}611 & 1.51 & 1.48 & 42\% \\
PRISM      & AdaEv.  & 65{,}437 & 26{,}713 & 2.45 & 2.23 & 35\% \\
Cloudcast  & OpenEv. & 23{,}600 & 24{,}937 & 0.95 & 1.27 & 35\% \\
Cloudcast  & EvoX    & 57{,}573 & 35{,}879 & 1.60 & 3.83 & 62\% \\
Cloudcast  & AdaEv.  & 59{,}361 & 38{,}649 & 1.54 & 2.35 & 41\% \\
Txn        & OpenEv. & 31{,}800 & 32{,}935 & 0.97 & 2.26 & 32\% \\
Txn        & EvoX    & 26{,}641 & 37{,}284 & 0.71 & 1.79 & 32\% \\
Txn        & AdaEv.  & 58{,}631 & 32{,}479 & 1.81 & 2.36 & 33\% \\
\bottomrule
\end{tabular}
\caption{Cost of one proposal on \texttt{gemini-2.5-flash}, beside the same nine cells on
\texttt{gpt-5-mini}. Calls and reasoning are the gemini figures.}
\label{tab:gemini}
\end{table*}

\begin{table}[t]
\centering\footnotesize
\setlength{\tabcolsep}{4.5pt}
\begin{tabular}{@{}llrrr@{}}
\toprule
& & \multicolumn{2}{c}{evolve / sampling} \\
\cmidrule(lr){3-4}
Task & Method & prop. & tok. & cut \\
\midrule
Cloudcast & AdaEv. & 1.61 & 1.59 & 3\% \\
Cloudcast & OpenEv. & 1.58 & 1.55 & 5\% \\
Cloudcast & EvoX & 1.53 & 1.53 & 0\% \\
Txn & EvoX & 1.25 & 1.25 & 0\% \\
Txn & AdaEv. & 1.27 & 1.25 & 6\% \\
PRISM & AdaEv. & 1.15 & 1.15 & 0\% \\
PRISM & EvoX & 1.15 & 1.15 & 0\% \\
Txn & OpenEv. & 1.16 & 1.15 & 7\% \\
PRISM & OpenEv. & 1.06 & 1.06 & 0\% \\
\midrule
mean & & 1.31 & 1.30 & \\
\bottomrule
\end{tabular}
\caption{Gemini evolution divided by gemini independent sampling, built as Table~\ref{tab:main} and
audited the same way.}
\label{tab:gemini2}
\end{table}

\begin{figure*}[t]
\centering
\includegraphics[width=\textwidth]{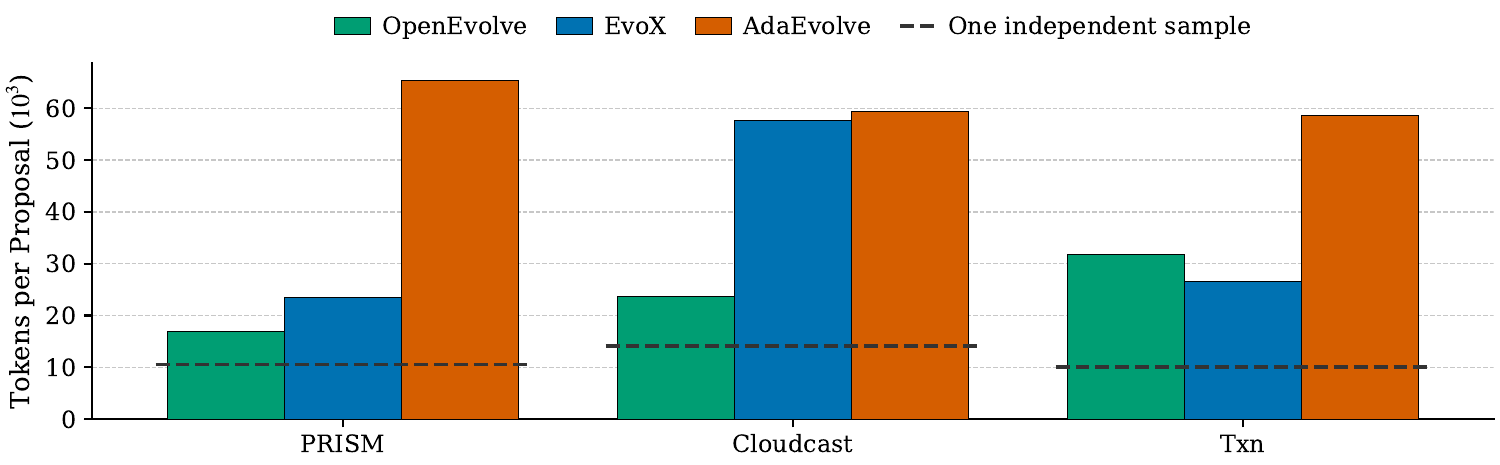}
\caption{Tokens for one proposal on \texttt{gemini-2.5-flash}, by method and task, with one
independent sample as the dashed line. The companion to Figure~\ref{fig:bars}.}
\label{fig:gemini_bars}
\end{figure*}

\begin{figure*}[t]
\centering
\includegraphics[width=\textwidth]{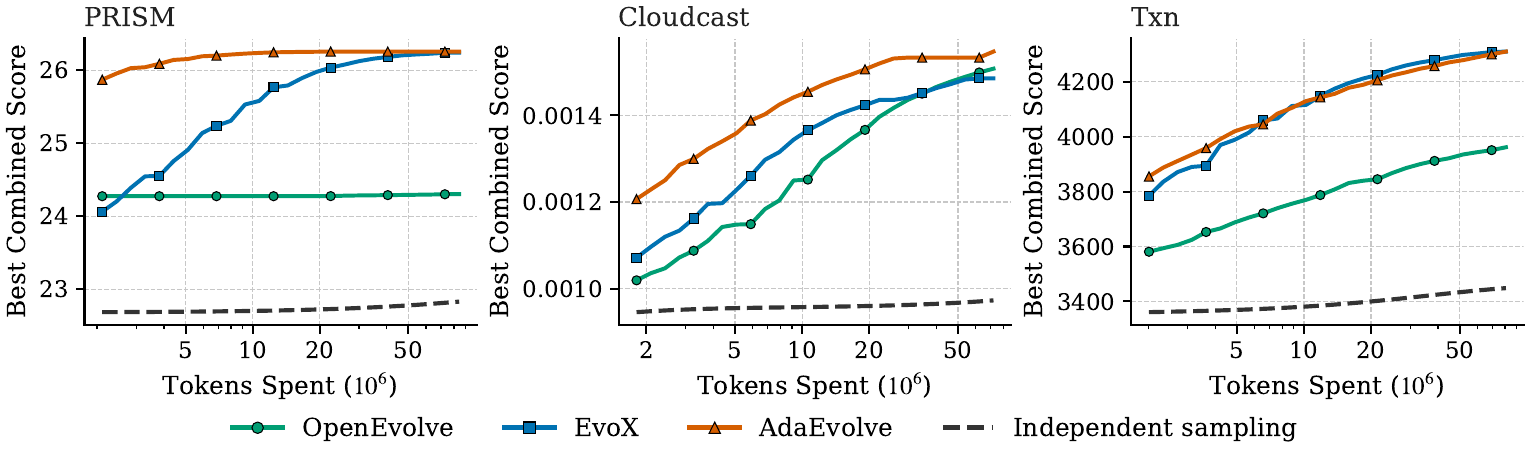}
\caption{Best score against tokens spent on \texttt{gemini-2.5-flash}, one panel per systems task,
with the same gates and audits as Figure~\ref{fig:curves}.}
\label{fig:gemini_curves}
\end{figure*}

\section{Confidence intervals}
\label{app:ci}

The intervals in Table~\ref{tab:ci} come from 2{,}000 replicates that resample both sides: a cell's
forty seed curves, with replacement, and the sampling pool, with replacement. The ratio is recomputed in
each replicate, so the interval covers the finite seed pool and the finite draw pool together. In three
cells the best seed count is forty, which makes the statistic a sample maximum, where the bootstrap is not
consistent; those use subsampling instead, $m=20$ of $40$ without replacement, whose spread is wider than
the true forty-seed spread and so cannot understate the interval. It moves one bound, Heilbronn with
OpenEvolve from $1.18$ to $1.17$. 

\begin{table}[t]
\centering\footnotesize
\setlength{\tabcolsep}{4.5pt}
\begin{tabular}{@{}llrr@{}}
\toprule
Task & Method & tok. & 95\% CI \\
\midrule
Heilbronn     & EvoX    & 1.28 & 1.24--1.36 \\
Heilbronn     & OpenEv. & 1.27 & 1.17--1.34 \\
Heilbronn     & AdaEv.  & 1.23 & 1.19--1.30 \\
Cloudcast     & AdaEv.  & 1.05 & 1.04--1.19 \\
Cloudcast     & EvoX    & 1.05 & 1.03--1.18 \\
Circle        & EvoX    & 1.04 & 1.04--1.04 \\
Circle        & AdaEv.  & 1.04 & 1.03--1.04 \\
Txn           & EvoX    & 1.04 & 1.03--1.06 \\
Txn           & OpenEv. & 1.02 & 1.02--1.05 \\
Txn           & AdaEv.  & 1.02 & 1.01--1.04 \\
Circle        & OpenEv. & 1.01 & 1.00--1.02 \\
PRISM         & AdaEv.  & 1.00 & 1.00--1.02 \\
PRISM         & EvoX    & 1.00 & 1.00--1.02 \\
PRISM         & OpenEv. & 1.00 & 1.00--1.02 \\
Cloudcast     & OpenEv. & 0.99 & 0.97--1.12 \\
\bottomrule
\end{tabular}
\caption{Equal-token ratios from Table~\ref{tab:main} with their $95\%$ intervals. PRISM's lower bounds are
$1.0009$ and round to $1.00$. Only Cloudcast with OpenEvolve spans parity.}
\label{tab:ci}
\end{table}

\section{Budgeting by model calls}
\label{app:calls}

\begin{figure*}[t]
\centering
\includegraphics[width=\textwidth]{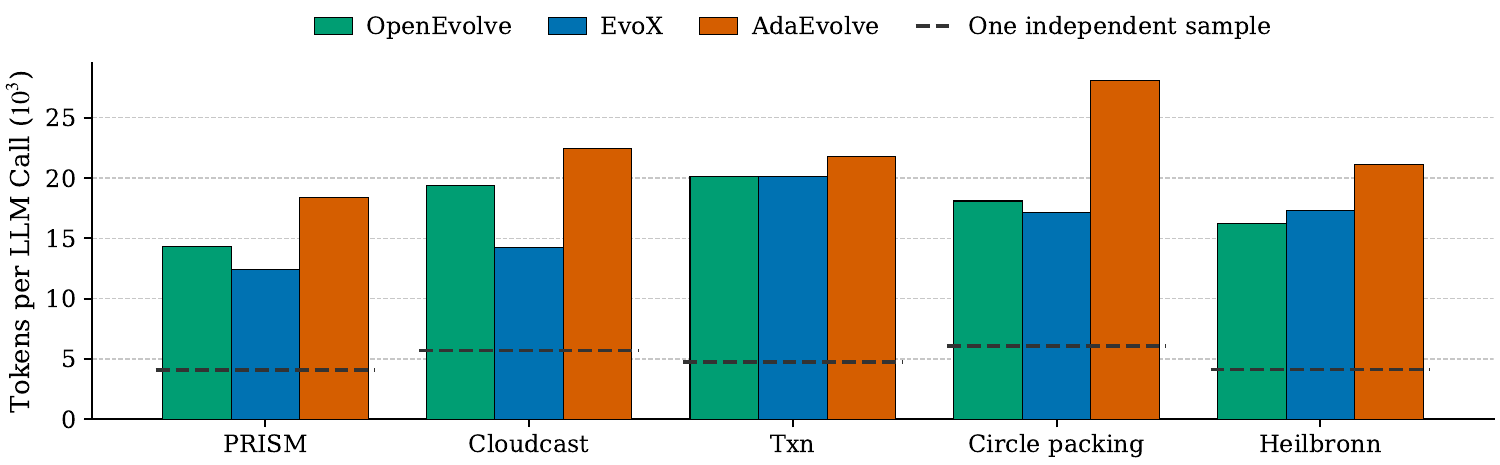}
\caption{Tokens for one model call, by method and task, with the cost of one independent sampling
call as the dashed line. If a call were a unit of compute this would be flat.}
\label{fig:percall}
\end{figure*}

Once proposals are shown not to be a unit of compute, the natural fallback is to count API calls.
That fails for the same reason: an evolve prompt carries the program being edited and part of the archive,
and both grow as the run proceeds, so a later call is a larger call. Figure~\ref{fig:percall} shows it
directly. Tokens per call run from 12.4k in the cheapest cell to 28.1k in the dearest, a $2.3\times$
spread against $2.5\times$ for tokens per proposal, so calls remove about a twelfth of the variation. One
evolve call costs $3.3$ to $4.4\times$ one sampling call, against $4.7$ to $7.2\times$ per proposal.

Table~\ref{tab:calls} repeats Table~\ref{tab:main} with a call budget beside the other two. On the
sampling side one call is one draw, so a budget of $B$ calls buys $B$ draws; on the evolve side the
cumulative call count replaces cumulative tokens. The mean ratio moves from $1.085$ at equal proposals to
$1.083$ at equal calls, against $1.069$ at equal tokens: calls recover about an eighth of what tokens
recover, and leave the same $11$ of $15$ pairs within $5\%$ where tokens leave $12$. The clearest case is
Cloudcast with OpenEvolve, reading $1.02$ at equal proposals, $1.01$ at equal calls and $0.99$ at equal
tokens. The one pair where sampling wins is visible only when the budget is counted in tokens.

\begin{table}[t]
\centering\footnotesize
\setlength{\tabcolsep}{4.5pt}
\begin{tabular}{@{}llrrr@{}}
\toprule
& & \multicolumn{3}{c}{evolve / sampling} \\
\cmidrule(lr){3-5}
Task & Method & prop. & calls & tok. \\
\midrule
Heilbronn     & EvoX    & 1.33 & 1.33 & 1.28 \\
Heilbronn     & OpenEv. & 1.31 & 1.31 & 1.27 \\
Heilbronn     & AdaEv.  & 1.32 & 1.29 & 1.23 \\
Cloudcast     & AdaEv.  & 1.06 & 1.06 & 1.05 \\
Cloudcast     & EvoX    & 1.05 & 1.05 & 1.05 \\
Txn           & EvoX    & 1.04 & 1.04 & 1.04 \\
Circle        & EvoX    & 1.04 & 1.04 & 1.04 \\
Circle        & AdaEv.  & 1.04 & 1.04 & 1.04 \\
Txn           & AdaEv.  & 1.03 & 1.03 & 1.02 \\
Txn           & OpenEv. & 1.03 & 1.03 & 1.02 \\
Circle        & OpenEv. & 1.02 & 1.02 & 1.01 \\
PRISM         & EvoX    & 1.00 & 1.00 & 1.00 \\
PRISM         & AdaEv.  & 1.00 & 1.00 & 1.00 \\
PRISM         & OpenEv. & 1.00 & 1.00 & 1.00 \\
Cloudcast     & OpenEv. & 1.02 & 1.01 & 0.99 \\
\midrule
mean & & 1.09 & 1.08 & 1.07 \\
\bottomrule
\end{tabular}
\caption{The same fifteen pairs at three budget definitions. Transaction scheduling and Cloudcast are
audited as in Table~\ref{tab:main}. Switching from proposals to calls changes almost nothing; switching to
tokens is what moves the comparison and what matters.}
\label{tab:calls}
\end{table}

\section{What each scorer accepts}
\label{app:audit}

A search is only as good as the scorer behind it. This appendix lists what each task's scorer will
accept from a program that has not done the job: a schedule covering 3 of 300 transactions, a broadcast
that drops partitions or misstates its own price, a server answering 2\% of requests. Table~\ref{tab:audit}
adds how often such a program was the one a practitioner would have shipped, which is the number that
matters, because selection takes the maximum rather than the average. Three of the four had to be caught by
re-running the program; only PRISM's is visible in the logs. Circle packing and Heilbronn are here as
controls, and their evaluators check their own output, so nothing turns up.

\begin{table}[t]
\centering\footnotesize
\setlength{\tabcolsep}{4pt}
\begin{tabular}{@{}llrl@{}}
\toprule
Task & Accepted despite & Degenerate & Effect \\
\midrule
Txn       & 3 of 300 placed        & 3/120  & $5.06\!\rightarrow\!1.02$ \\
Cloudcast & no partitions sent     & 2/113  & $<0.1\%$ \\
Cloudcast & price misreported      & 2/113  & $1.08\!\rightarrow\!1.05$ \\
PRISM     & 2\% of requests served & 22/120 & gated out \\
Circle    & nothing                & 0/120  & control \\
Heilbronn & nothing                & 0/120  & control \\
\bottomrule
\end{tabular}
\caption{What each scorer accepts, how often it happens, and what removing it costs. The Degenerate column
counts seeds whose own best program fails a direct coverage check, out of the seeds we re-executed;
both Cloudcast rows rest on the same 113, because seven of its winners cannot be re-run at all. Circle
packing and Heilbronn accept nothing because their evaluators validate the output themselves, the latter
rejecting any result that does not place every point inside the triangle. The Effect column is the change to
Table~\ref{tab:main}: transaction scheduling loses its headline, Cloudcast is unaffected by the two that
send no partitions because neither ever led, but one of the two that misreport their price was a cell
ceiling, and PRISM's gate excludes its own before they are scored.}
\label{tab:audit}
\end{table}

\end{document}